# Meta-information Guided Cross-domain Synergistic Diffusion Model for Low-dose PET Reconstruction

Mengxiao Geng, Ran Hong, Xiaoling Xu, Bingxuan Li, Qiegen Liu, *Senior Member, IEEE*

*Abstract*—Low-dose PET imaging is crucial for reducing patient radiation exposure but faces challenges like noise interference, reduced contrast, and difficulty in preserving physiological details. Existing methods often neglect both projection-domain physics knowledge and patient-specific meta-information, which are critical for functional-semantic correlation mining. In this study, we introduce a meta-information guided cross-domain synergistic diffusion model (MiG-DM) that integrates comprehensive cross-modal priors to generate high-quality PET images. Specifically, a meta-information encoding module transforms clinical parameters into semantic prompts by considering patient characteristics, dose-related information, and semi-quantitative parameters, enabling cross-modal alignment between textual meta-information and image reconstruction. Additionally, the cross-domain architecture combines projection-domain and image-domain processing. In the projection domain, a specialized sinogram adapter captures global physical structures through convolution operations equivalent to global image-domain filtering. Experiments on the UDPET public dataset and clinical datasets with varying dose levels demonstrate that MiG-DM outperforms state-of-the-art methods in enhancing PET image quality and preserving physiological details.

## I. INTRODUCTION

Positron emission tomography (PET) is a non-invasive nuclear imaging technique that visualizes metabolic activities like glucose uptake by detecting positron annihilation events, making it crucial for early disease diagnosis and treatment monitoring in oncology, neurology, and cardiology [1]-[3]. However, PET imaging faces significant challenges. Its functional nature results in images with inherently fuzzy boundaries and low spatial resolution. This is mainly due to the physical principles of positron emission, such as positron range and detector resolution [4]. Furthermore, the quality of PET images is highly dependent on the radiation dose administered. While low-dose protocols can reduce patient radiation risk, they also lead to increased noise levels and reduced image quality, which in turn compromise diagnostic accuracy [5]. Additionally, PET imaging relies on radioactive tracers with short half-lives, such as $^{18}$F-FDG. The production and handling of these tracers require specialized equipment and strict safety protocols, which limits the accessibility and widespread use of PET imaging [6].

With the development of artificial intelligence, deep learning-based approaches have shown promise in improving PET image quality, leveraging convolutional neural networks (CNNs) [8]-[10] or generative adversarial networks (GANs) [11]-[13] to map noisy input data to high-quality images [14]. For example, Peng *et al*. [15] proposed a novel PET image reconstruction method by integrating CT images as inputs into a 3D U-Net-based [16] architecture to boost image quality. Chen *et al*. [17] introduced a 3D image space shuffle U-Net, incorporating shuffle/unshuffled layers into the U-Net architecture for low-dose reconstruction. Yang *et al*. [18] proposed a conditional weakly-supervised multi-task learning strategy with a multi-channel self-attention module, which improves noise reduction and contrast recovery by incorporating an auxiliary task as an anatomical regularize. Ouyang *et al*. [19] employed a basic GAN with task-specific perceptual loss for PET reconstruction through adversarial learning [20]-[22], and incorporated a pre-trained Amyloid classifier for guidance.

Recently, diffusion models have emerged as a powerful class of generative models, capable of learning complex data distributions and generating high-fidelity samples [23]-[24]. For example, Gong *et al*. [25] applied denoising diffusion probabilistic models (DDPM) [26] to PET image reconstruction, integrating MR prior information and PET data-consistency constraints to enhance performance and reduce uncertainty. Shen *et al*. [27] proposed the bidirectional condition diffusion probabilistic model, which learns a score function network [28] via evidence lower-bound optimization and employs two handcrafted conditions in latent space to generate high-quality images. Jiang *et al*. [29] created an unsupervised PET enhancement framework using a latent diffusion model trained on full-dose PET images, incorporating PET compression, Poisson diffusion, and CT-guided cross-attention. Han *et al*. [30] proposed a diffusion model-based PET reconstruction framework with a coarse prediction module and an iterative refinement module, enhanced by auxiliary guidance and contrastive diffusion strategies. Moreover, Huang *et al*. [31] developed a diffusion-transformer model integrating diffusion and transformer techniques with a joint compact prior to boost image quality and protect lesion details. Pan *et al*. [32] designed a diffusion-based PET consistency model that enhances low-dose PET image quality by learning a consistency function during reverse diffusion and employing shifted windows as visual transform-

This work was supported in part by the National Key Research and Development Program of China under Grants 2023YFF1204300 and 2023YFF1204302, the National Natural Science Foundation of China under Grant 62122033, and the Key Research and Development Program of Jiangxi Province under Grant 20212BBE53001. (M. Geng and R. Hong are co-first authors.) (Corresponding authors: B. Li and Q. Liu.)

M. Geng, R. Hong, X. Xu and Q. Liu are with School of Information Engineering, Nanchang University, Nanchang 330031, China. (mxiaogeng @163.com, ranhong@email.ncu.edu.cn, xuxiaoling@ncu.edu.cn, liuqiegen@ncu.edu.cn)

B. Li is with Anhui Province Key Laboratory of Biomedical Imaging and Intelligent Processing, Institute of Artificial Intelligence, Hefei Comprehensive National Science Center, Hefei 230088, China. (libingxuan@iai.ustc.edu.cn)

ers. Xie *et al.* [33] established a dose-aware diffusion model for 3D low-dose PET imaging using neighboring slices as conditional information. Nevertheless, most of them typically operate in a single domain, either the image domain or the sinogram domain, without fully exploiting the complementary information between these two domains. Furthermore, these methods predominantly focus on image-domain features (e.g., texture, intensity distributions), neglecting the complementary value of meta-information. This oversight limits their ability to exploit functional-semantic correlations, leading to suboptimal preservation of physiological details in reconstructed images.

Considering all the above factors, we propose a **Me**ta-**i**nformation **G**uided cross-domain synergistic **D**iffusion **M**odel (MiG-DM) for low-dose PET reconstruction. This model bridges the gap by aligning textual semantic cues with image reconstruction, thereby enhancing both structural fidelity and functional interpretability. At the same time, the cross-domain framework, which jointly processes data in the image and sinogram domains, offers a more comprehensive understanding of the PET imaging process, enabling more effective noise suppression and feature preservation. The main contributions of this work are summarized as follows:

● *Adaptive MI Encoding for Functional-semantic Deep Coupling in PET Reconstruction.* A meta-information (MI) encoding module is introduced to achieve cross-modal alignment of MI semantics with image reconstruction in PET imaging. By converting PET-specific functional parameters into semantic prompts, considering patient characteristics, dose-related information, and semi-quantitative parameters, the module facilitates the creation of semantic prompt vectors through an MI encoder.

● *Cross-domain Synergy for Global Physical and Local Detail Optimization.* A reconstruction framework that integrates projection-domain and image-domain processing is proposed to utilize both global and local information. In the projection domain, there is a specialized sinogram adapter that transforms raw projection data into feature space, effectively capturing the global physical structure of radiation distribution.

The remainder of this paper is organized as follows. Section II provides a concise overview of related works in the field. Section III elaborates on the key idea of the proposed method. The experimental setup and corresponding results are detailed in Section IV. A comprehensive discussion of the findings is presented in Section V, and the paper concludes with a succinct summary in Section VI.

## II. PRELIMINARY

### A. Diffusion Models

Diffusion models have shown great potential for PET image reconstruction, where the primary goal is to restore high-quality images from extremely noisy data. It can typically be divided into a forward diffusion process and a reverse denoising process. The forward diffusion process is a continuous-time stochastic process that acts on a full-dose PET image $x_0$, continuously injecting noise as the time step increases until the image satisfies pure Gaussian noise. The forward diffusion process can be represented by a Markov chain:

$$q(x_{1:T} | x_0) = \prod_{t=1}^{T} q(x_t | x_{t-1}) \quad (1)$$

$$q(x_t | x_{t-1}) = \mathcal{N}(x_t | \sqrt{\alpha_t} x_{t-1}, \beta_t I) \quad (2)$$

where $\alpha_t = 1 - \beta_t$ and $\bar{\alpha}_t = \prod_{i=1}^{t} \alpha_i$. Here, $\beta_t$ represents the variance schedule that controls the amount of noise added.

The reconstruction process reverses the forward process by learning to predict the noise at each time step. The reverse denoising process is also a continuous-time stochastic process and can be represented by a Markov chain:

$$p(x_{t-1} | x_t) = \mathcal{N}(x_{t-1} | \mu_t(x_t), \sigma_t^2 I) \quad (3)$$

$$\mu_t(x_t) = \frac{1}{\sqrt{\alpha_t}}(x_t - \frac{\beta_t}{\sqrt{1-\bar{\alpha}_t}} \varepsilon_\theta(x_t, t)),\ \sigma_t = \sqrt{\frac{1-\bar{\alpha}_{t-1}}{1-\bar{\alpha}_t} \beta_t} \quad (4)$$

where $\mu_t$ and $\sigma_t^2$ represent the mean and variance of the Gaussian distribution, respectively. $\sigma_t^2$ is a known and fixed parameter, while $\mu_t$ needs to be learned and predicted through a neural network. Its optimization objective is expressed as:

$$\min_{\theta} \mathbb{E}_{t, x_0, \varepsilon} \| \varepsilon - \varepsilon_\theta(x_t, t) \|_2^2 \quad (5)$$

where $\varepsilon$ denotes the noise added in the diffusion process, and $\varepsilon_\theta$ represents the predicted noise. Since the diffusion model learns image features at different levels of noise, it can learn the image distribution more efficiently and achieve more outstanding generation quality compared to other generative models. Moreover, by injecting conditions into the reverse denoising process, the diffusion model can achieve accurate generation control and complete conditional generation tasks such as Image-to-image [34],[35] or text-to-image [36],[37].

### B. LoRA

For fine-tuning large pre-trained models, full parameter updates incur substantial computational and memory overhead. To address this, low-rank adaptation (LoRA) [38] introduces an efficient decomposition strategy that freezes the original pre-trained weights and learns task-specific updates through two low-rank matrices $B \in \mathbb{R}^{r \times d}$ and $A \in \mathbb{R}^{d \times r}$, where the rank $r \ll d$ (typically $r \in \{4, 8, 16\}$). The effectiveness of LoRA is grounded in the low intrinsic dimensionality of neural networks [39]. During the full fine-tuning, meaningful parameter updates primarily occur within a low-dimensional subspace. The low-rank projection of LoRA effectively captures these essential updates, achieving comparable performance to full fine-tuning with dramatically reduced resource requirements.

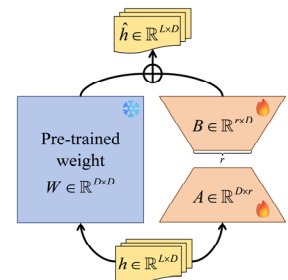

Fig. 1. LoRA fine-tuning via frozen pretrained weights and low-rank updates.

## III. METHOD

This section details the MiG-DM framework that synergizes meta-information guidance with cross-domain diffusion modelling for low-dose PET reconstruction.

### A. Motivation

Meta-information in PET imaging encompasses patient-specific physiological parameters and metrics, including body height, weight, radiotracer injection dose, and dynamic imaging parameters. These data encode functional and physiological contexts critical for interpreting metabolic activity visualized in PET images. Unlike anatomical imaging modalities like CT and MRI, PET relies on functional signals where meta-information correlates with biological processes such as glucose metabolism or receptor binding. For instance, Halpern *et al.* [40] demonstrated that overweight patients require an extended PET acquisition protocol. Masuda *et al.* [41] evaluated the effect of optimizing injected dose on the image quality of overweight patients using lutetium oxyorthosilicate PET/CT with high-performance detector electronics. Gu *et al.* [42] explored the diagnostic image quality and lesion detectability of BMI-based reduced injection doses for $^{18}$F-FDG PET/CT imaging. The relevance of meta-information to PET reconstruction is twofold: (1) it provides quantitative links between imaging signals and physiological reality, enabling more accurate modeling of radiotracer distribution; (2) it serves as prior knowledge to constrain the reconstruction process, particularly in low-dose scenarios where noise and artifacts compromise image quality. However, conventional deep learning-based methods mainly focus on image-domain features such as texture and intensity distributions, neglecting the complementary value of meta-information. This oversight limits their ability to exploit functional-semantic correlations, leading to suboptimal preservation of physiological details in reconstructed images.

Moreover, the sinogram represents the raw projection data in PET imaging, encoding the spatial distribution of coincidence events from radioactive tracers across multiple angular views. Employing projection-domain data for low-dose PET reconstruction offers several key advantages over image-domain methods. First, the projection domain inherently preserves the physical constraints of radiation transport, such as ray trajectories and emission distributions, thereby enabling direct modeling of the underlying physics. Second, unlike image-domain methods that are confined to local convolutional operations as illustrated in **Fig. 2**, the projection domain facilitates global information processing. Local operations in the projection domain induce global effects in the image domain, making cross-domain reconstruction particularly effective in optimizing both fine-scale and large-scale features for improved fidelity. Third, the sinogram exhibits structured redundancy derived from its geometric relationship between angular and radial coordinates. This structure enables efficient signal extraction without the need to contend with complex spatial textures present in image-domain data, allowing models to focus on genuine signal patterns. Finally, the projection domain is compatible with physical priors such as attenuation maps and system response functions. This compatibility helps mitigate challenges posed by limited data in low-dose scenarios, enhancing reconstruction fidelity by integrating domain-specific knowledge.

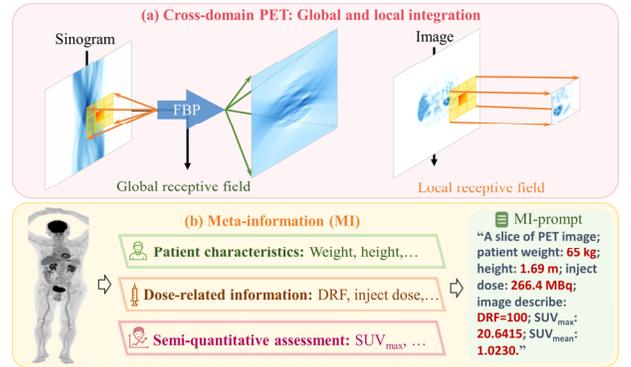

**Fig. 2.** Meta-information guided cross-domain synergistic diffusion model. (a) Cross-domain PET reconstruction integrating global and local features. (b) Meta-information components and their use in reconstruction.

Therefore, we develop a meta-information guided cross-domain synergistic diffusion model. By integrating a meta-information encoding module with a cross-domain architecture, the proposed model bridges the gap between textual metadata and image reconstruction while optimizing the use of cross-modal prior information. **Fig. 3** illustrates the overall procedure of the proposed method.

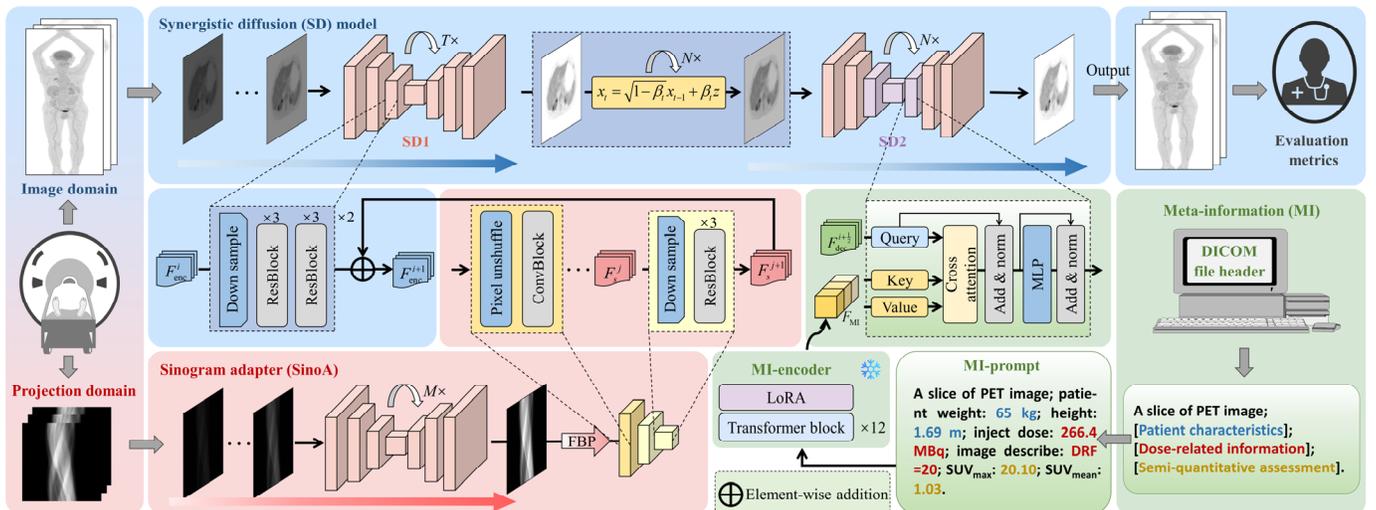

**Fig. 3.** Overview of the MiG-DM framework. The MI-encoder sustains functional-semantic representations through cross-modal alignment of meta-information and image. The cross-domain reconstruction diffusion model ensures physical consistency by learning and integrating information from both the image and projection domains.

## B. Adaptive MI Encoding Guidance

In the MiG-DM framework, the incorporation of multi-modal information plays a pivotal role. Subsequently, the adaptive multi-modal encoding guidance process is elaborated through three key components: MI-prompt preparation, MI-encoder design, and contrastive learning.

*MI-prompt Preparation:* Achieving precise alignment between meta-information and PET images is crucial, and the formulation of text prompts plays a pivotal role in this process. However, the formulation of such text prompts faces two main challenges. Firstly, conventional text prompts are typically designed for natural images [43], focusing on object categories, thus failing to capture the complex metabolic information of human anatomy contained in PET images. Secondly, text prompts generated by visual-language models [44] often concentrate on visual aspects such as color and shape, insufficient to deliver the in-depth functional information required for PET imaging. To address these challenges, we introduce a MI-prompt that incorporates effective functional-semantic information. As depicted in **Fig. 4**, the MI-prompt is derived from headers of DICOM-format PET images and categorized into three token components including [Patient characteristics], [Dose-related information], and [Semi-quantitative assessment]. The [Patient characteristics] component covers patient basics like height and weight, offering the model prior information on body size and basal metabolism. The [Dose-related information] component includes details of radiation intensity like radiotracer dosage and low-dose levels, providing prior information on image radiation intensity. The [Semi-quantitative assessment] component presents semi-quantitative parameters like $SUV_{max}$ and $SUV_{mean}$, enabling the model to gauge pathological information and cancer progression from each PET image slice.

*MI-encoder Design:* Inspired by the vision-language model CLIP [45], we design a dedicated learning framework to obtain an MI encoded feature $F_{MI}$ that is aligned with PET images. First, we pre-train on large-scale paired natural images and texts to enable the model to initially acquire image-text understanding capabilities and semantic alignment between images and descriptive texts. In the fine-tuning process, the pre-trained parameters are copied and frozen as the initialization for the fine-tuning model. The model is then continuously trained on paired data of PET images with multiple dosage levels and MI-prompt. This design allows the model to retain its basic image-text understanding while further learning and mastering MI comprehension abilities.

**Fig. 4** presents the detailed architecture of the MI-encoder. The image encoder accepts a PET image $x \in \mathbb{R}^{C \times H \times W}$, where $C$, $H$, and $W$ represent the channel, height, and width of the image, respectively. After undergoing a patch-embedding operation, the image is transformed into a serialized vector $F_{img} \in \mathbb{R}^{n \times D}$, with $n$ being the number of patches and $D$ the dimension of the image feature vector. The image-encoder is composed of 12 stacked ViT blocks [46]. The intermediate features of $F_{img}$ within the image encoder can be expressed as $\{F_{img}^0, \cdots, F_{img}^i, \cdots, F_{img}^{12}\}$. Correspondingly, the MI-encoder receives a MI-prompt $m \in \mathbb{R}^L$, where $L$ represents the length of the MI-prompt. After passing through the tokenizer layer, it is converted into a serialized vector $F_{MI} \in \mathbb{R}^{L \times D}$. The MI-encoder is composed of 12 stacked Transformer blocks [47]. The intermediate representations of $F_{MI}$ input into the MI-encoder are $\{F_{MI}^0, \cdots, F_{MI}^i, \cdots, F_{MI}^{12}\}$. In the fine-tuning process, a trainable LoRA module [48] is added to each ViT block and Transformer block. In the $i+1$-th Transformer block of the MI-encoder, the multi-head attention module operates by modifying the query, key, and value via integration with the LoRA module. The modified query $\bar{Q}$, key $\bar{K}$, and value $\bar{V}$ are computed as follows:

$$\bar{Q} = W_Q \times F_{MI}^i + LoRA_Q(F_{MI}^i) = W_Q \times F_{MI}^i + B_Q A_Q \times F_{MI}^i$$
$$\bar{K} = W_K \times F_{MI}^i + LoRA_K(F_{MI}^i) = W_K \times F_{MI}^i + B_K A_K \times F_{MI}^i \quad (6)$$
$$\bar{V} = W_V \times F_{MI}^i + LoRA_V(F_{MI}^i) = W_V \times F_{MI}^i + B_V A_V \times F_{MI}^i$$

where $W_Q, W_K, W_V \in \mathbb{R}^{D \times D}$ represent the pre-trained weights. The output of the multi-head attention module is derived as

$$F_{MI}^{i+\frac{1}{2}} = \text{Softmax}(\bar{Q} \cdot \bar{K}^T / \sqrt{D}) \cdot \bar{V} \quad (7)$$

Additionally, $A \in \mathbb{R}^{r \times D}$, $B \in \mathbb{R}^{D \times r}$, and $r$ represents the rank of the LoRA weights. The smaller $r$ is, and the smaller the calculation amount increased by fine-tuning is. In the pre-training stage, $A$ is initialized with Kaiming initialization while $B$ performs zero initialization. This is to smoothly add LoRA weights as the fine-tuning process proceeds and prevent the gradient change from being too drastic. A similar approach is applied to add LoRA weights to the weight matrix of the multilayer perceptron (MLP):

$$F_{MI}^{i+1} = W_{MLP} \times F_{MI}^{i+\frac{1}{2}} + B_{MLP} A_{MLP} \times F_{MI}^{i+\frac{1}{2}} \quad (8)$$

The MLP consists of two linear layers with the Gaussian error linear unit (GeLU) activation function. Thus, the weight matrices of MLP and LoRA can be expressed as $W_{MLP} = \{W_0, W_1\}$ and $B_{MLP} A_{MLP} = \{B_{L_0} A_{L_0}, B_{L_1} A_{L_1}\} = \{LoRA_{L_0}, LoRA_{L_1}\}$.

*Contrastive Learning:* In the pre-training and fine-tuning processes, contrastive learning is utilized to align image and MI-prompt embeddings by maximizing the similarity for correct pairings and minimizing it for incorrect ones. Specifically, the image encoder transforms PET image data into feature representations $F_{img}^{12} \in \mathbb{R}^{n \times D}$, which are then subjected to global average pooling and $L_2$ normalization, resulting in $\hat{F}_{img} \in \mathbb{R}^D$. At the same time, the MI-encoder converts MI features $F_{MI}^{12} \in \mathbb{R}^{r \times D}$ into $\hat{F}_{MI} \in \mathbb{R}^D$ after passing through a linear projection layer and $L_2$ normalization. Once the PET image features and MI features are mapped into the same feature space, a contrastive loss function $\mathcal{L}_{Con}$ between the image and MI features is optimized to align the two modalities below:

$$\mathcal{L}_{Con} = -\frac{1}{2D}\{\sum_{i=1}^{D}\log(\frac{\exp(\hat{F}'_{MI}\hat{F}_{img}/\tau)}{\sum_{j=1}^{D}\exp(\hat{F}'_{MI}\hat{F}_{img}/\tau)})$$
$$+ \sum_{i=1}^{D}\log(\frac{\exp(\hat{F}'_{img}\hat{F}_{MI}/\tau)}{\sum_{j=1}^{D}\exp(\hat{F}'_{img}\hat{F}_{MI}/\tau)})\} \quad (9)$$

where $\tau$ represents learnable temperature parameter that regulates the contrastive strength. The loss function optimizes the model in both directions: from MI-prompt to PET image and from PET image to MI-prompt. It compels the model to align similar semantics while amplifying the semantic disparities between distinct features.

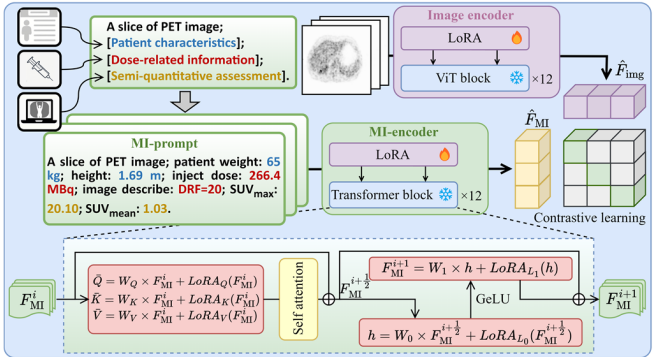

Fig. 4. Contrastive learning framework of MI-encoder with integrated patient characteristics, dose-related information, and semi-quantitative assessment derived from DICOM PET images.

### C. Cross-domain Synergistic Diffusion Reconstruction

In this subsection, we first introduce a sinogram adapter (SinoA) designed to learn the distribution of physical information in the projection domain. This information is then transferred to a synergistic diffusion (SD) model consisting of two components, SD1 and SD2. SD1 integrates the projection-domain features derived from SinoA into the image-domain reconstruction process, while SD2 incorporates MI-encoded features to enhance the functional semantics.

*Sinogram Adapter:* The sinogram reconstruction model (SRM), based on diffusion principles, is initially utilized to restore the integrity of low-dose projection data. The model adopts a U-Net framework with an encoder-decoder structure. The encoder consists of six hierarchical feature extraction stages, each containing six residual blocks (ResBlock). The first stage maintains the original input resolution, and subsequent stages progressively down-sample the feature maps to resolutions of 1/2, 1/4, 1/8, 1/16, and 1/32. The decoder employs a symmetric architecture with skip connections to enable multi-scale feature fusion. Next, a cross-domain synergistic mapper (CDSM) connects the projection and image domains. The module processes the original input $F_s = R(y_0)$, where $y_0$ denotes the output sinogram of the SRM and $R$ represents the traditional reconstruction algorithm [49]. The transformation begins with a pixel unshuffling operation with a scale factor of 4, followed by a convolution block (ConvBlock) to generate the features $F_s^0$. These features are then refined through four processing stages, each comprising three residual blocks [50]. Stages 2 to 4 include additional down-sampling operations with a scale factor of 2, producing multi-scale feature representations $\{F_s^1, F_s^2, F_s^3, F_s^4\}$ at resolutions of 1/4, 1/8, 1/16, and 1/32 of the original input dimensions.

*Synergistic Diffusion Model:* As the key generation process of MiG-DM, the synergistic diffusion model combines SD1 for projection-domain feature integration with SD2 for meta-information incorporation, achieving high-quality PET image reconstruction with preserved functional semantics. The specific architectures of SD1/SD2 and their connection mode are elaborated below.

To begin with, SD1 and SRM share the same model structure. Recognizing the critical role of the encoder in image understanding and the decoder in feature reconstruction, we inject projection-domain features from SinoA into the encoder of SD1 to enhance both local and global feature comprehension. The final four stages of the SD1 encoder generate feature maps $\{F_{enc}^1, F_{enc}^2, F_{enc}^3, F_{enc}^4\}$ at resolutions of 1/4, 1/8, 1/16, and 1/32. Since these features share resolutions with the projection-domain feature group, their fusion employs element-wise addition as follows:

$$\hat{F}_{enc}^i = F_{enc}^i + F_s^i, i \in \{1,2,3,4\} \tag{10}$$

Notably, projection-domain guidance is restricted to the final four encoder stages of SD1 because parameters in deeper layers converge markedly faster than those in shallow counterparts. Consequently, the deep features can rapidly adapt to external conditioning, whereas shallow layers retain stable, data-driven representations of generic textures and edges. By injecting conditional information only at these deeper levels, we enhance semantic consistency and fine-detail fidelity without unduly perturbing the low-level feature hierarchy.

Then, SD2 and SRM share similar architectures, with the key distinction being the incorporation of cross attention modules [51] in the final three stages of both encoder and decoder, as well as intermediate layers. When processing feature maps $F_{dec}^i$ at the $i$-th layer of the SD2 decoder, the output $F_{dec}^{i+\frac{1}{2}}$ from the ResBlock interacts with the MI feature $F_{MI}$ through cross attention:

$$Q = W_Q \times F_{dec}^{i+\frac{1}{2}}, \ K = W_K \times F_{MI}, \ V = W_V \times F_{MI} \tag{11}$$

$$\text{CrossAttn} = \text{Softmax}(\frac{Q \cdot K^T}{\sqrt{D}}) \cdot V \tag{12}$$

where the image feature $F_{dec}^{i+\frac{1}{2}}$ serves as Query while the MI feature $F_{MI}$ provides both Key and Value.

Finally, we implement a resample strategy [52] to connect SD1 and SD2. Specifically, the reconstructed image $\hat{x}$ from SD1 undergoes $N$-timestep diffusion:

$$\hat{x}_N = \sqrt{\overline{\alpha}_N} \hat{x} + \sqrt{1-\overline{\alpha}_N} \varepsilon \tag{13}$$

where $\varepsilon \sim \mathcal{N}(0, I)$ being a sample from the Gaussian distribution. SD2 then performs reverse reconstruction from timestep $N$ to obtain the high quality PET image. It is worth noting that SD1 is injected with image-domain and projection-domain, thus endowing it with enhanced physical structure reconstruct capabilities. In contrast, SD2 is injected with MI-prompts, thereby granting it stronger detail and semantic reconstruct capabilities. The resampling step $N$ can serve as a hyperparameter to balance the structural and semantic quality. **Theorem 1** demonstrates the existence of an appropriate $N$ enables the optimization of the reconstructed image quality across the two models. In our experiments, we set $N = 50$. An ablation study and further discussion of this choice are provided in **Section V**.

*Theorem 1 (Existence of an Optimal Resample Timestep).* Consider a resampling process involving SD1 and SD2 over a total of $T$ diffusion steps. Let $N \in [0,T]$ and $Q(N)$ denote the number of resampling timesteps and the image quality metric, respectively. If the following conditions hold:

(i) SD1 excels over SD2 in structural generation, $S_1 > S_2$.

(ii) SD2 excels over SD1 in semantic generation, $D_2 > D_1$.

(iii) The decay constants satisfy $k_S > 0$, $k_D > 0$, $k_S \neq k_D$.

Then there exists at least one optimal resampling timestep $N^* \in [0,T]$ that maximizes $Q(N)$. Moreover, if the addition condition $\frac{(S_1 - S_2)k_S}{(D_2 - D_1)k_D} > 0$ is satisfied and the quantity $\frac{1}{k_S - k_D} \ln(\frac{(S_1 - S_2)k_S}{(D_2 - D_1)k_D})$ lies in the open interval $(0,T)$, then the optimal solution $N^*$ is given explicitly by

$$N^* = \frac{1}{k_S - k_D} \ln(\frac{(S_1 - S_2)k_S}{(D_2 - D_1)k_D}) \tag{14}$$

The proof is provided in **Appendix B**.

In summary, the MiG-DM algorithm is provided for low-dose PET reconstruction, as seen in **Algorithm 1**.

---

**Algorithm 1: MiG-DM**

**Iterative reconstruction**

**1: Input:** schedule $\alpha$ and $\beta$, noise $\varepsilon \sim \mathcal{N}(0,I)$, MI-prompt $m$

**2:** $F_{MI} = \text{MI-encoder}(m)$

**3: For** $t = M - 1$ **to** 0 **do**

**4:** $\hat{y}_{t-1} = \frac{1}{\sqrt{\alpha_t}}(\hat{y}_t - \frac{\beta_t}{\sqrt{1 - \bar{\alpha}_t}}\text{SRM}(\hat{y}_t, t)) + \sqrt{\frac{1 - \bar{\alpha}_{t-1}}{1 - \bar{\alpha}_t}}\beta_t \varepsilon$

**5: End for**

**6:** $\{F_s^1, F_s^2, F_s^3, F_s^4\} = \text{CDSM}(R(y_0))$

**7: For** $t = T - 1$ **to** 0 **do**

**8:** $\hat{x}_{t-1} = \frac{1}{\sqrt{\alpha_t}}(\hat{x}_t - \frac{\beta_t}{\sqrt{1 - \bar{\alpha}_t}}\text{SD1}(\hat{x}_t, t, \{F_s^1, F_s^2, F_s^3, F_s^4\}))$
$+ \sqrt{\frac{1 - \bar{\alpha}_{t-1}}{1 - \bar{\alpha}_t}}\beta_t \varepsilon$

**9: End for**

**10: Resample** $\hat{x}_N = \sqrt{\bar{\alpha}_N}\hat{x}_0 + \sqrt{1 - \bar{\alpha}_N}\varepsilon$

**11: For** $t = N - 1$ **to** 0 **do**

**12:** $\hat{x}_{t-1} = \frac{1}{\sqrt{\alpha_t}}(\hat{x}_t - \frac{\beta_t}{\sqrt{1 - \bar{\alpha}_t}}\text{SD2}(\hat{x}_t, t, F_{MI})) + \sqrt{\frac{1 - \bar{\alpha}_{t-1}}{1 - \bar{\alpha}_t}}\beta_t \varepsilon$

**13: End for**

**14: Output:** $x_0$

---

### D. Impact of Repetition Patterns in Synergistic Diffusion Model

The SinoA-guided SD1 model and the MI-guided SD2 model play crucial roles. However, the repetition pattern between them is a key factor influencing the reconstruction quality. Therefore, we design two different repetition patterns including "SD1-to-SD2 cascade" and "SD2-to-SD1 cascade" in **Fig. 5**. For the "SD1-to-SD2 cascade" mode, the SinoA-guided SD1 is performed first, followed by the MI-guided SD2 reconstruction via resample. For the "SD2-to-SD1 cascade" mode, the MI-guided SD2 reconstruction is carried out first, and then the SinoA-guided SD1 reconstruction is performed. Because meta-information guidance mainly aims to adjust the functional semantics of the PET image, we ultimately select the "SD1-to-SD2 cascade" repetition pattern. If reconstruction is directly performed without a certain basic morphology, the basic morphology of the results will be degraded. Although the SinoA-guided SD1 model will be run afterwards to optimize the reconstructed morphology, it will cause the functional semantics to degrade. Hence, the second pattern is less favorable than the first one.

Moreover, we choose to split and inject the image-domain PET images, projection-domain sinograms, and MI-prompt into the diffusion model as three separate components. A dedicated network focusing on single-modal data facilitates fast convergence and strong robustness, while the reconstruction probability distribution obtained by separating individual conditions is equivalent to that derived from aggregating all conditions. Relevant theoretical analysis detail is included in **Theorem 2**, while the specific experimental results are documented in **Section V**.

*Theorem 2 (Multi-Condition Guided Reverse Reconstruction).* Let $lq$ denote a low-dose PET image, $y$ a sinogram, and $m$ meta-information. Assume that three conditions $\{lq, y, m\}$ are conditionally independent given $x_t$. Then the score of the multi-conditional reverse transition can be decomposed as

$$\begin{aligned}&\nabla_{x_t} \log p(x_t \mid x_{t+1}, lq, y, m) \\ &= \nabla_{x_t} \log p(x_t \mid x_{t+1}) + \sum_{c \in \{lq, y, m\}} \lambda_c \nabla_{x_t} \log p(c \mid x_t)\end{aligned} \tag{15}$$

where $\lambda_c > 0$ are scaling coefficients that control the relative strength of guidance from each condition.

Proof is deferred to **Appendix C**.

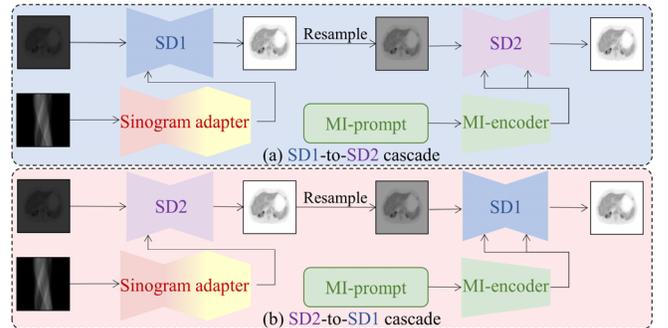

**Fig. 5.** Different repetition patterns in synergistic diffusion model.

## IV. EXPERIMENTS

### A. Experimental Setup

In this section, the performance of MiG-DM is compared with state-of-the-art methods, including U-Net [52], MPRNet [16], ViT-Rec [46], Pix2Pix [53], and IDDPM [24]. To ensure comparability and fairness of the experiments, all methods are

7 IEEE xxxxxxxxx, VOL. xx, NO. x, 2025

conducted on the same datasets. Open-source code is available at: https://github.com/yqx7150 /MiG-DM.

*Datasets:* Two distinct datasets were employed to conduct a comprehensive evaluation. The first was the *UDPET* dataset obtained from the MICCAI 2024 ultra-low-dose PET imaging challenge, which comprises low-dose PET and full-dose PET images with dose reduction factors (DRFs) of 10 and 20. Low-dose images were generated by subsampling full scans and were perfectly aligned with their full-dose PET images. Each patient case included 673 axial 2D slices, cropped to 256×256 to remove background. All imaging data were acquired using the uEXPLORER whole-body PET system for $^{18}$F-FDG-PET applications. The training cohort consisted of 101 patients per dose level, providing 67,973 slices for model development, while an independent set of 1,346 slices was reserved for testing. Secondly, the *Clinical* dataset selected from 10 patients was utilized to further assess the generalization capability of these models. Each patient case contained 450 axial 2D image slices, zero-padded to 256×256 for testing. The data was sampled from the DigitMI 930 PET/CT scanner that developed by RAYSOLUTION Healthcare Co., Ltd. The scanner integrates fully digital photon detectors and offers an axial field of view of 30.6 cm. Each scan covered 4 to 8 beds, with scan times ranging from 45 seconds to 3 minutes per bed. Low-dose PET data was obtained by resampling the listmode data into random intervals, retaining the random data per cycle and discarding the remaining data.

*Parameter Configuration:* For the fine-tuning of the MI-encoder, we set the LoRA rank $r = 4$ and trained for 1000 iterations on paired PET images and MI-prompts using a batch size of 256 and a learning rate of $1\times10^{-4}$ with the Adam optimizer. For the SD1, SD2, and SRM models, SD1 was trained on full-dose PET data, SD2 was trained on full-dose data and corresponding MI-prompts, and SRM was trained on paired low-dose and full-dose sinograms. These three models utilized a batch size of 8 and a learning rate of $1\times10^{-4}$ with the AdamW optimizer, training for 300,000 iterations. Moreover, we used DDSA to connect the fine-tuned SD1 and SRM, and trained on paired low-dose sinograms and full-dose PET images for 100,000 iterations with a batch size of 8 and a learning rate of $1\times10^{-4}$ using the AdamW optimizer. All training and testing experiments were conducted using two NVIDIA GeForce RTX 3090 GPUs, each with 24 GB of memory.

*Performance Evaluation:* To quantitatively measure the error caused by MiG-DM, the peak signal-to-noise ratio (PSNR), structural similarity (SSIM), mean squared error (MSE) and learned perceptual image patch similarity (LPIPS) [54] are used to evaluate the quality of reconstruction images. To further evaluate the performance on the *Clinical* dataset, we have additionally incorporated clinical metrics which includes the difference of the maximum standardized uptake value ($\triangle$SUV$_{max}$) and the difference of the mean standardized uptake value ($\triangle$SUV$_{mean}$), both represent the quantitative gap between the reconstructed results and reference for the lesion, as well as the tumor-to-background (TBR) and contrast ratio (CR). The specific expressions for TBR and CR are as follows:

$$\text{TBR} = \text{SUV}_{\text{mean lesion}} / \text{SUV}_{\text{mean liver}} \quad (16)$$

$$\text{CR} = \text{SUV}_{\text{max lesion}} / \text{SUV}_{\text{mean liver}} \quad (17)$$

where SUV$_{\text{mean lesion}}$ and SUV$_{\text{mean liver}}$ represent the mean standardized uptake value for the lesion and liver, respectively. SUV$_{\text{max lesion}}$ is the maximum standardized uptake value for the lesion.

### B. Reconstruction Experiments

*Comparison on UDPET Public Dataset:* To assess the efficacy of MiG-DM using the *UDPET* dataset, Table I displays a comprehensive quantitative analysis across different DRFs and patient weight groups. The results demonstrate that MiG-DM consistently achieves higher PSNR and SSIM values, while exhibiting lower MSE and LPIPS values compared to other reconstruction methods. Specifically, for the ≤60 kg weight group at a DRF of 20, MiG-DM attains a PSNR vlaue of 46.15 dB and an SSIM value of 0.9884, which are among the highest scores recorded. Moreover, MiG-DM outperforms the second-best IDDPM by 0.94 dB in PSNR, 0.0013 in SSIM, 0.6617 in MSE, and 0.0053 in LPIPS. These improvements underscore the superior ability of MiG-DM to maintain image quality and detail preservation, especially under higher dose reduction factors, thereby enhancing the overall diagnostic value of low-dose PET imaging.

TABLE I
COMPARISON OF STATE-OF-THE-ART METHODS IN TERMS OF AVERAGE PSNR ↑ , SSIM ↑ , MSE (*E-3) ↓ , AND LPIPS ↓ UNDER VARIOUS DRFS ON THE *UDPET* DATASET. ↓ REPRESENTS THE SMALLER THE BETTER, AND ↑ REPRESENTS THE BIGGER THE BETTER. THE **BOLD** AND *ITALIC* FONTS INDICATE THE OPTIMAL AND SUB-OPTIMAL VALUES, RESPECTIVELY.

| *UDPET* dataset | ≤60 kg | | ≥80 kg | |
|---|---|---|---|---|
| | DRF=10 | DRF=20 | DRF=10 | DRF=20 |
| U-Net | 33.68/0.9848/0.8902/0.0228 | 32.11/0.9803/1.1464/0.0286 | 32.97/0.9776/1.1612/0.0313 | 31.09/0.9723/1.5241/0.0390 |
| MPRNet | 34.28/0.9848/0.5250/0.0231 | 30.41/0.9115/2.4648/0.1559 | 34.07/0.9808/0.5678/0.0275 | 29.72/0.8782/1.4827/0.1767 |
| ViT-Rec | 32.59/0.8956/0.7790/0.1278 | 30.20/0.8964/1.3453/0.1655 | 32.60/0.9123/0.7042/0.1106 | 30.10/0.9064/1.3426/0.1508 |
| Pix2Pix | 43.83/0.9807/1.3951/0.0394 | 45.25/0.9881/0.7759/0.0248 | 43.75/0.9831/1.1507/0.0359 | 42.47/0.9800/1.2321/0.0403 |
| IDDPM | 45.84/0.9886/0.8664/**0.0196** | 45.21/0.9871/0.8971/0.0237 | 43.79/0.9833/1.1685/0.0271 | 42.98/0.9809/1.2186/0.0290 |
| MiG-DM | **46.46/0.9895/0.1824**/*0.0225* | **46.15/0.9884/0.2354/0.0184** | **44.40/0.9852/0.2624/0.0225** | **43.61/0.9820/0.3708/0.0284** |

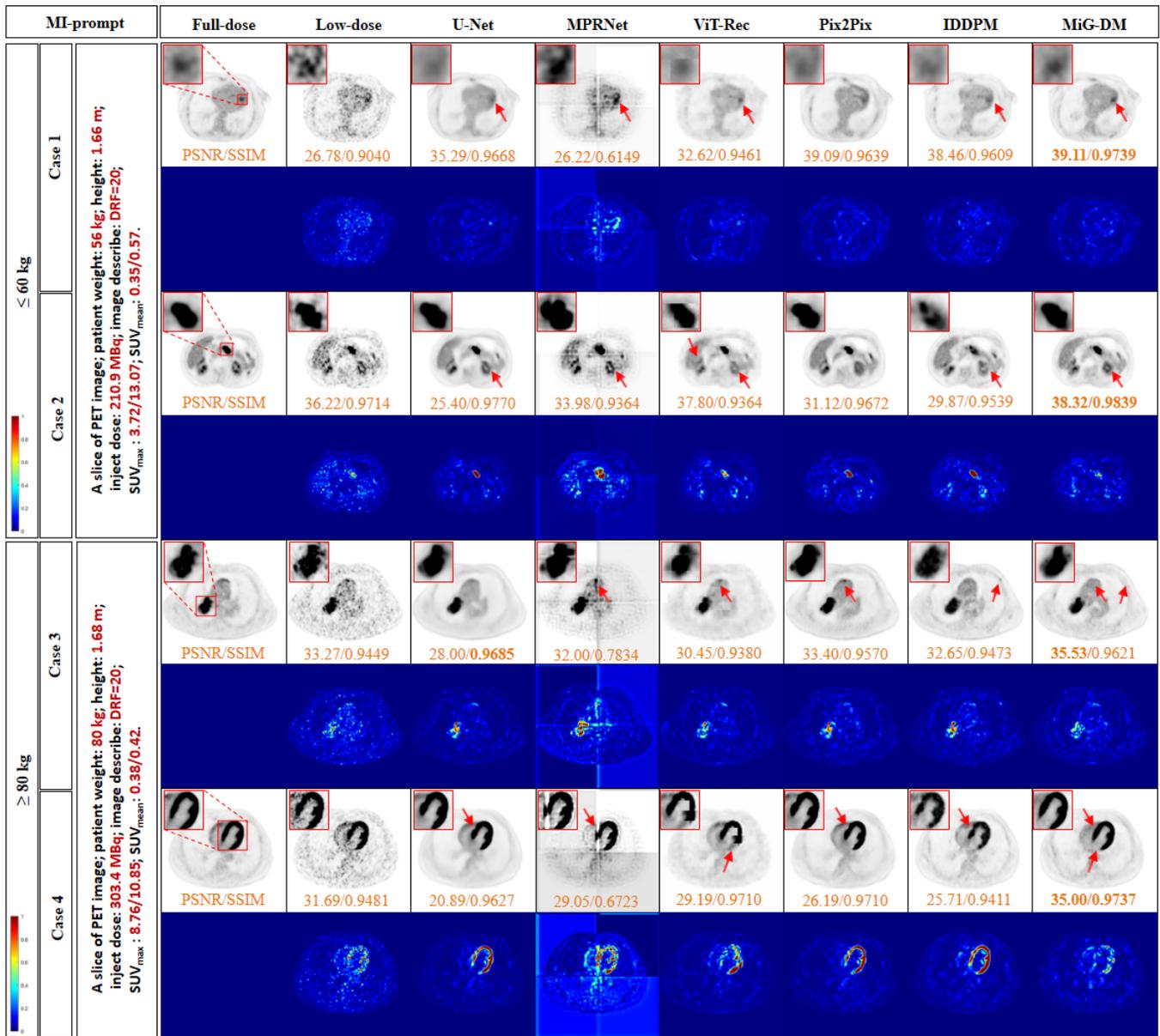

**Fig. 6.** Reconstruction results on the *UDPET* public dataset for different weight patients at DRF=20. From left to right: Full-dose, Low-dose, reconstruction by UNet, MPRNet, ViT-Rec, Pix2Pix, IDDPM, and MiG-DM (Ours). The red box shows an enlarged view of the ROI region, while the corresponding error map is presented below each reconstructed PET image.

Fig. 6 visually illustrates the reconstruction outcomes of different methods on the *UDPET* public dataset for patients of varying weights at DRF=20. Reconstruction results by MiG-DM are closest to the full-dose PET images, effectively suppressing noise while retaining physiological details. For instance, in case 1, the ROI marked by the red box contains a positive tumor lesion. Compared with the full-dose image, MiG-DM exhibits superior accuracy in terms of detailed information such as lesion morphology, clarity and quantitative distribution. In contrast, other methods including U-Net, ViT-Rec, Pix2Pix and IDDPM all show degradation and blurring of lesion morphology as well as inaccuracies in quantitative information within the ROI. The error maps illustrate the differences between reconstructed images and full-dose images. In case 4, the patient's cardiac region is presented, with the arrows pointing to the peripheral contour of the heart. Compared with the full-dose image, the contour reconstructed by MiG-DM is distinct and smooth, the other methods exhibit shape defects of varying degrees on the left side and below the heart. In summary, the visual results demonstrate that the integration of local and global information within the cross-domain reconstruction framework enables the reconstructed images to achieve favorable performance in both overall quality and local details. Moreover, the MI-encoder endows MiG-DM with functional semantics, thereby ensuring the effectiveness of functional reconstruction in ROIs.

To more accurately evaluate the agreement between reconstructed images and full-dose images, as shown in Fig.7, we present the reconstructed images of various methods and their corresponding Bland-Altman plots, which depict the numerical differences between each method's reconstructed images and the full-dose images. Where the red line represents the mean of the differences, and the blue dashed lines indicate the 95% limits of agreement (LoA). The results demonstrate that the images reconstructed by MiG-DM achieve the lowest mean difference of -0.0018 and the narrowest 95% LoA, ranging from -0.0360 to 0.0324. Meanwhile, the red arrows in the reconstructed images point to the detailed regions. For other methods, reconstruction errors in these detailed regions lead to the presence of extreme values in their Bland-Altman plots and an expansion of the 95% LoA range. Therefore, the smaller mean difference and 95% LoA of MiG-DM indicate that it has distinct advantages in both global image quality and local detailed lesions.

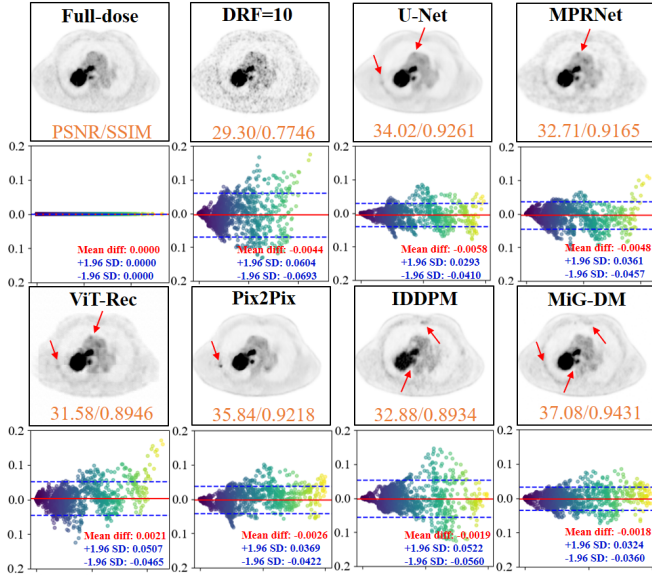

**Fig. 7.** Bland-Altman plots and reconstruction performance on the *UDPET* dataset at DRF=10. Red line represents mean difference between reconstructed and full-dose images, while blue lines denote 95% limits of agreement.

In the left half of Fig. 8, the maximum intensity projection (MIP) image of the patient and the reconstruction results of MiG-DM in the coronal, sagittal and axial view under DRF of 10 and 20 are presented. MiG-DM obtains superior results from multiple perspectives. For instance, the red boxes in the coronal and sagittal highlight the detailed reconstruction of the patient's brain, where MiG-DM successfully reconstructs the accurate morphology of white matter and gray matter under both DRF=10 and DRF=20. Additionally, the bule boxes in the coronal and sagittal, as well as the red box in the axial view, mark the reconstruction results of the patient's abdominal tumor, which exhibit a morphology close to that of the full-dose image under both DRF conditions. In the right half of Fig. 8, box plots of PSNR and SSIM metrics for various methods are presented. At both DRF=10 and DRF=20, MiG-DM has the highest median values, the smallest interquartile range and contains fewer extreme values. In contrast, other SOTA methods, even if they exhibit a smaller interquartile range, tend to have lower metrics and a greater number of extreme values.

*Comparison on Clinical Dataset:* To evaluate the performance of different methods under clinical conditions, we tested various approaches using the *Clinical* dataset. Fig. 9 presents the coronal reconstruction results and corresponding error maps of U-Net, MPRNet, ViT-Rec, Pix2Pix, IDDPM and MiG-DM, where the red arrows indicate key lesion locations. For MiG-DM, the value of error map is generally low, and with accurate reconstruction of both shapes and quantitative values of multiple lesions. In contrast, although the reconstruction results of U-Net and IDDPM are visually acceptable, their error maps reveal deficiencies in preserving functional numerical fidelity. Meanwhile, MPRNet, ViT-Rec and Pix2Pix all exhibit blurring and distortion at lesion locations. In summary, MiG-DM remains capable of ensuring reconstruction performance in terms of visual presentation, complex structures and functional semantics simultaneously in clinical assessment.

To quantitatively evaluate the performance of SOTA methods on *Clinical* dataset, we employed image quality metrics and PET clinical metrics. Table II reports the image quality metrics of different methods, where MiG-DM achieved the highest PSNR of 34.98 dB and SSIM of 0.9284, as well as the lowest MSE of 5.4466(*E-3) and LPIPS of 0.0516. In the *Clinical* dataset, MiG-DM demonstrated greater advantages over other methods. Conversely, the performance of some methods, such as U-Net and Pix2Pix, degraded significantly. This indicates that MiG-DM possesses stronger generalization ability and robustness.

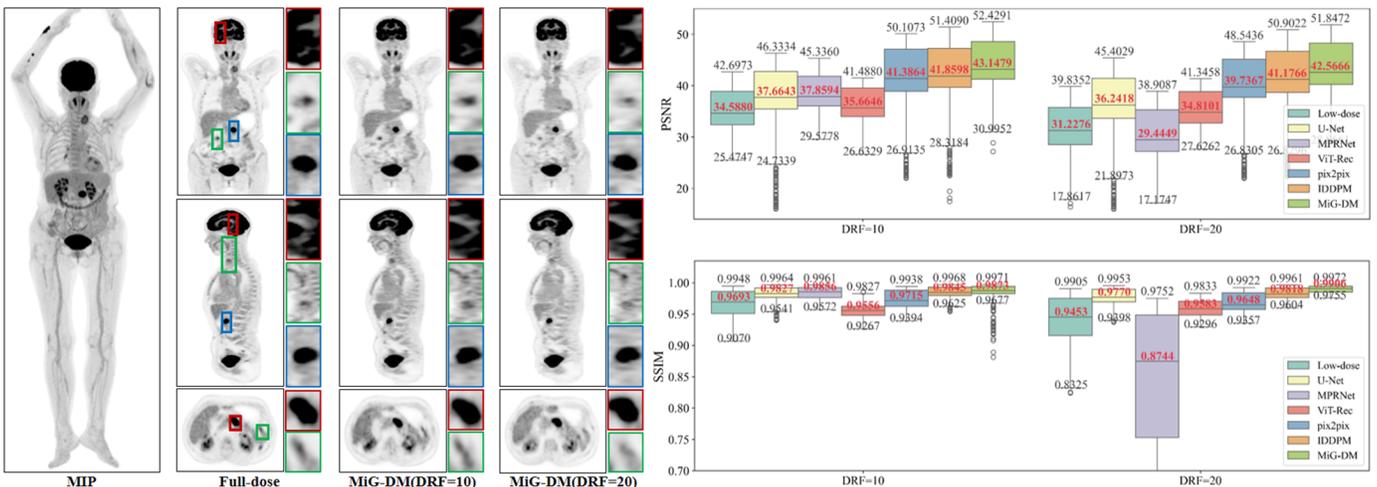

**Fig. 8.** Reconstruction results of MiG-DM on the *UDPET* dataset, along with visualizations of quantitative evaluation metrics for different methods. From left to right: MIP image, full-dose reference, and reconstruction results of MiG-DM at DRF=10 and DRF=20. Box plots illustrate the comparisons of PSNR and SSIM across multiple methods under DRF=10 and DRF=20.

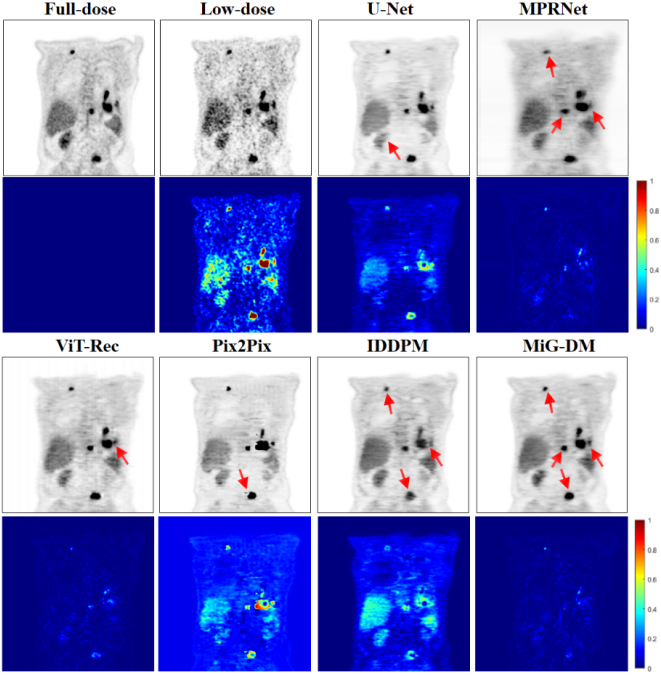

**Fig. 9.** Coronal views of reconstruction results on the *Clinical* dataset. From: Full-dose, Low-dose, reconstruction by UNet, MPRNet, ViT-Rec, Pix2Pix, IDDPM and MiG-DM (Ours).

TABLE II
COMPARISON OF STATE-OF-THE-ART METHODS IN TERMS OF THE AVERAGE PSNR, SSIM, MSE(*E-3), AND LPIPS ON THE *CLINICAL* DATASET.

| *Clinical* dataset | PSNR↑ | SSIM↑ | MSE↓ | LPIPS↓ |
|---|---|---|---|---|
| U-Net | 23.45 | 0.9049 | 6.2150 | 0.1206 |
| MPRNet | 28.91 | 0.5967 | 6.1386 | 0.1081 |
| ViT-Rec | 30.76 | 0.9197 | 7.6829 | 0.1134 |
| Pix2Pix | 28.49 | 0.9021 | 17.5988 | 0.1123 |
| IDDPM | 31.12 | 0.9248 | 15.9463 | 0.0812 |
| **MiG-DM** | **34.97** | **0.9284** | **5.4466** | **0.0516** |

Table III compares the results of PET clinical metrics, including average $\Delta SUV_{max}$, $\Delta SUV_{mean}$, TBR and CR. Smaller $\Delta SUV_{max}$ and $\Delta SUV_{mean}$ indicate a smaller discrepancy in metabolic indices compared with the reference full-dose images. Higher TBR denotes a greater contrast between the lesion and the background, rendering the lesion more distinct. Similarly, higher CR signifies a stronger contrast between the lesion and the liver, meaning the lesion is more prominent. The results of MiG-DM demonstrate that it attains the lowest $\Delta SUV_{max}$ of 1.4751 and $\Delta SUV_{mean}$ of 0.0404, which significantly outperform other methods, for example, IDDPM yields a $\Delta SUV_{max}$ of 2.9282 and $\Delta SUV_{mean}$ of 0.0848. This highlights that MiG-DM exhibits precise metabolic fidelity preservation capability. Additionally, in comparison with other methods, MiG-DM gains the highest TBE of 0.6416 and CR of 2.0652. These metrics underscore the superiority and efficacy of MiG-DM in enhancing tissue homogeneity and lesion conspicuity. In conclusion, the cross-domain reconstruction framework ensures that the model maintains superior reconstruction performance across different datasets. Meanwhile, the MI-encoder injects functional semantics into the model, endowing it with reliability in preserving the clinical functional information of PET.

TABLE III
COMPARISON OF STATE-OF-THE-ART METHODS.

| *Clinical* dataset | $\Delta SUV_{max}$↓ | $\Delta SUV_{mean}$↓ | TBR↑ | CR↑ |
|---|---|---|---|---|
| U-Net | 3.0160 | 0.1262 | 0.2614 | 0.9171 |
| MPRNet | 2.9524 | 0.0849 | 0.4201 | 0.8176 |
| ViT-Rec | 2.6124 | 0.0618 | 0.5825 | 1.2037 |
| Pix2Pix | 2.9328 | 0.1103 | 0.3641 | 1.0272 |
| IDDPM | 2.9282 | 0.0848 | 0.4107 | 0.8177 |
| **MiG-DM** | **1.4751** | **0.0404** | **0.6416** | **2.0652** |

### C. Ablation Study

*Impact of MI-encoder and SinoA:* To assess the influence of the MI-encoder and SinoA modules within MiG-DM, an ablation study was conducted on the *UDPET* dataset at DRF=20, with results presented in Table IV. We adopted the model excluding the MI-encoder and SinoA modules as the baseline, which obtained PSNR of 45.21, SSIM of 0.9871, MSE of 0.8971(*E-3) and LPIPS of 0.0237. With the introduction of the MI-encoder module, the MSE decreased significantly from 0.8971(*E-3) to 0.2388(*E-3), which indicates that the functional semantic information introduced by the MI-encoder enhances the model's capability in restoring the quantitative values of lesion regions. When only the SinoA module is introduced, PSNR and MSE exhibit slight degradation, while SSIM and LPIPS show marginal improvement. This indicates that the global information introduced by SinoA enhances the overall reconstruction performance, such as global contrast, statistical consistency, and perceptual fidelity. Thus, when both MI-encoder and SinoA modules are integrated, the model synergistically leverages global statistic and functional semantic information, thereby enhancing reconstruction performance in both dimensions.

TABLE IV
QUANTITATIVE RESULTS OF ABLATION STUDY ON MI-ENCODER AND SINOA.

| MI-encoder | SinoA | PSNR↑ | SSIM↑ | MSE↓ | LPIPS↓ |
|---|---|---|---|---|---|
| ✗ | ✗ | 45.21 | 0.9871 | 0.8971 | 0.0237 |
| ✓ | ✗ | 45.94 | 0.9878 | 0.2388 | 0.0202 |
| ✗ | ✓ | 45.04 | 0.9874 | 0.8985 | 0.0216 |
| ✓ | ✓ | **46.15** | **0.9884** | **0.2354** | **0.0184** |

*Impact of LoRA and MI-prompt:* To evaluate the fine-tuning effect of the LoRA module, we conducted an ablation experiment where the MI-encoder with and without the LoRA module was incorporated into the cross-domain reconstruction framework, and their image quality metrics were displayed in Table V. Compared with the MI-encoder without LoRA, the incorporation of LoRA into the MI-encoder resulted in performance gains of 0.41, 0.001, 0.0052(*E-3), and 0.0118 in PSNR, SSIM, MSE and LPIPS, respectively. The improvement in LPIPS is particularly significant, indicating that the fine-tuned MI-encoder can extract more effective MI representations. Such high-quality MI representations contribute substantially to enhancing the model's perception performance and image quality.

TABLE V
QUANTITATIVE RESULTS OF ABLATION STUDY ON LORA.

| DRF=20 | PSNR↑ | SSIM↑ | MSE↓ | LPIPS↓ |
|---|---|---|---|---|
| MI-encoder w/o LoRA | 45.74 | 0.9874 | 0.2406 | 0.0302 |
| **MI-encoder w/ LoRA** | **46.15** | **0.9884** | **0.2354** | **0.0184** |

Fig.10 visualizes the alignment probability distributions of the MI-encoder with and without LoRA under low-dose images at DRF=10, along with 1 correct MI-prompt and 8 incorrect MI-prompts. Without LoRA, the alignment probability for each MI-prompt is roughly identical, indicating that the MI-encoder without LoRA fails to identify the MI-prompt that correctly pairs with the PET image. In contrast, when equipped with LoRA, the MI-encoder can align the image with the correct MI-prompt with a probability of 89.06%. This demonstrates that LoRA enables the MI-encoder to acquire robust MI encoding capabilities through minimal parameter modifications. Furthermore, the MI-prompt consists of three components: Patient characteristics, Dose-related information and Semi-quantitative assessment. As illustrated in Fig. 10, any error in one of these components leads to a significant drop in the prediction probability, which implies that the three categories of information in the MI-prompt are equally important for the expression of functional semantics.

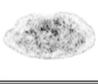

**Fig. 10.** Probability distributions of alignment for MI-Encoder with and without LoRA across various MI-prompts.

## V. DISCUSSION

We have demonstrated that the cross-domain reconstruction framework can effectively learn both global and local information of PET images, and the introduction of the MI-prompt enables the model to acquire valid functional semantic information. It is worth noting that the order of cross-domain information and MI-prompt injection, the number of resampling steps, along with the number of sampling steps of SRM in projection-domain, have a certain impact on the accuracy and reliability of image reconstruction.

*Analysis of Combination Pattern:* Table VI presents the performance of the two combination patterns of MiG-DM on the *UDPET* dataset. Specifically, the SD1-to-SD2 cascade first acquires global information through SD1, then obtains functional semantic via SD2. In contrast, the SD2-to-SD1 cascade alternates the injection order of the two types of guiding information. SD1-to-SD2 cascade outperforms SD2-to-SD1 cascade in terms of PSNR, SSIM and LPIPS. This indicates that under the SD1-to-SD2 cascade combination pattern, the model first fuses global and local visual information to obtain an initial visual reconstruction result. Subsequently, functional semantic information performs numerically precise adjustments on specific regions of the morphologically intact image, thereby ensuring the consistency of functional semantics. However, when the processing order is reversed, functional information is adjusted first, and the subsequent injection of global information from the projection domain undermines the previously injected functional information, leading to degradation in image reconstruction performance.

TABLE VI
PERFORMANCE COMPARISON OF DIFFERENT COMBINATION PATTERNS UNDER DRF=20 ON THE *UDPET* DATASET.

| Pattern | PSNR↑ | SSIM↑ | MSE↓ | LPIPS↓ |
|---|---|---|---|---|
| **SD1-to-SD2 cascade** | **46.15** | **0.9884** | 0.2354 | **0.0184** |
| **SD2-to-SD1 cascade** | 45.74 | 0.9881 | **0.2327** | 0.0190 |

*Analysis of Resample Timestep:* Resample is employed to connect SD1 and SD2, with the aim of adding a certain degree of noise to the high quality image distribution generated by SD1, thereby bringing it closer to the denoising trajectory of SD2. Subsequently, functional semantics from the MI-prompt are injected during the denoising process of SD2 to reconstruct the result. Table VII show a comparison of quantitative results when the resample timestep $N$ is set to 5, 40, 50, 75 and 150. The results indicate that when $N$=50, the highest PSNR of 46.15 and the maximum SSIM of 0.9884 are achieved. An insufficient number of resample timestep would leave inadequate timestep for the injection of functional semantics, while an excessive number of resample timestep would lead to significant loss of the basic reconstruction results, both scenarios result in a degradation of reconstruction quality.

TABLE VII
QUANTITATIVE RESULTS OF DIFFERENT RESAMPLE TIMESTEPS UNDER DRF=20 ON THE *CLINICAL* DATASET.

| Resample | $N$=5 | $N$=40 | $N$=50 | $N$=75 | $N$=150 |
|---|---|---|---|---|---|
| PSNR↑ | 45.79 | 46.07 | **46.15** | 46.06 | 45.83 |
| SSIM↑ | 0.9878 | 0.9884 | **0.9884** | 0.9884 | 0.9882 |

*Analysis of SRM Sampling Step:* In the projection domain, sinograms comprise numerous sinusoidal curves with varying amplitudes and phases, most of which exhibit similar shapes and complex textures. Consequently, the number of sampling step affects the generation of projection-domain sinograms to a certain extent, which in turn influences the quality of global guiding information, ultimately leading to a degradation in image reconstruction quality. Therefore, we conduct an analysis and discussion on the number of sampling steps in the SRM. Fig. 11 illustrates the sinograms generated by SRM at sampling steps of 250, 500, 750 and 1000. When the number of sampling steps is low, abnormal projection lines appear in the sinograms, and the reconstructed images of MiG-DM are affected by erroneous global information, resulting in blurred images with lost details. As the number of sampling steps increases, the quality of sinogram generation improves significantly. For example, at the position circled by the orange dashed line, the previously missing projection lines are completely supplemented by when sample step is set to 1000, and the corresponding results also gain accurate lesion contours.

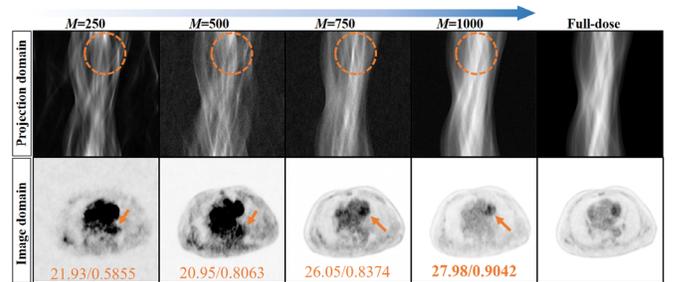

**Fig. 11.** Sinograms and corresponding reconstruction results of MiG-DM under different numbers of sampling steps.

## VI. CONCLUSION

In this study, we proposed a meta-information-guided cross-domain reconstruction framework. This framework simultaneously learns local and global information in PET images through the cross-domain reconstruction architecture, thereby generating results with preserved global structures from sinogram and maintained local details from PET images. Subsequently, the MI-encoder is used to extract mete-information from MI-prompt, infusing functional semantics into the model. This enables the model to enhance the accuracy of quantitative value prediction based on high quality morphological images. Our experiments have achieved performance that outperforms SOTA methods on both the public *UDPET* dataset and *Clinical* dataset. Future research directions may include further research directions may include refinement of model architecture, utilization of more diverse datasets and efforts to further enhance model performance and clinical applicability.

## APPENDIX

### A. Preliminary: Diffusion Model

Let $x_t$ denote the degraded image from pure image $x_0$ at timestep $t$, where $t = \{1, ..., T\}$. The forward diffusion process is defined by the Markov chain

$$q(x_{t+1}|x_t) = \mathcal{N}(x_{t+1} | \sqrt{\alpha_{t+1}} x_t, \beta_{t+1} I), \ t = 1, ..., T \quad (A.1)$$

with $\alpha_t = 1 - \beta_t$ and $\{\beta_t\}_{t=1}^T$ a predefined variance schedule. Define $\bar{\alpha}_t = \prod_{s=1}^t \alpha_s$. Then the distribution of $x_t$ given $x_0$ can be written as

$$q(x_t|x_0) = \mathcal{N}(x_t | \sqrt{\bar{\alpha}_t} x_0, (1-\bar{\alpha}_t) I) \quad (A.2)$$

The reverse reconstruction process aims to recover $x_0$ starting from $x_T \sim \mathcal{N}(0, I)$ by learning a parameterized Gaussian transition

$$p_\theta(x_{t-1}|x_t) = \mathcal{N}(x_{t-1}; \mu_\theta(x_t, t), \sigma_t^2 I) \quad (A.3)$$

where the mean $\mu_\theta(x_t, t)$ is predicted via a noise-prediction network $\varepsilon_\theta(x_t, t)$ as

$$\mu_\theta(x_t, t) = \frac{1}{\sqrt{\alpha_t}} (x_t - \frac{\beta_t}{\sqrt{1-\bar{\alpha}_t}} \varepsilon_\theta(x_t, t)) \quad (A.4)$$

### B. Optimality Analysis of Cascade Sampling

The cascade sampling process, which chains the output of SD1 to the input of SD2, requires careful tuning of the denoising process. The number of resampling steps $N$, is a critical hyper-parameter. The following analysis provides a theoretical characterization of its optimality.

**Lemma 1 (Joint Distribution of Cascade Sampling).** Let $\theta_1$ and $\theta_2$ denote the parameters of SD1 and SD2, respectively. $N \in \{1, \cdots, T\}$ is the intermediate resampling timestep. The cascade sampling process can be decomposed into three stages:

- The initial reconstructed sample $\hat{x}_0$ is generated using SD1. From Eq. (A.3), the reverse process is executed starting from $x_T \sim \mathcal{N}(0, I)$:

$$x_{t-1} \sim p_{\theta_1}(x_{t-1}|x_t), \ t = T, ..., 1. \quad (B.1)$$

to gain $\hat{x}_0$. This distribution is denoted as $p_1(\hat{x}_0)$.

- Perform $N$ steps forward noising on $\hat{x}_0$. From Eq. (A.2), we can explicitly compute $x_N$:

$$x_N = q(x_N | \hat{x}_0) = \mathcal{N}(x_N | \sqrt{\bar{\alpha}_N} \hat{x}_0, \sqrt{1-\bar{\alpha}_N} I) \quad (B.2)$$

- SD2 is utilized to perform iterative reconstruction starting from $x_N$, yielding $x_0$. Starting from $t = N$, the reverse reconstruction process is executed iteratively:

$$x_{t-1} \sim p_{\theta_2}(x_{t-1}|x_t), \ t = N, ..., 1. \quad (B.3)$$

to obtain the final sample $x_0$. Denote the distribution of $x_0$ given $x_N$ under this reverse process by $p_2(x_0|x_N)$.

Consequently, the joint distribution of cascade sampling process can be formulated as follows:

$$p(\hat{x}_0, x_N, x_0) = p_1(\hat{x}_0) q(x_N | \hat{x}_0) p_2(x_0 | x_N) \quad (B.4)$$

The marginal distribution of the final sample $x_0$ is given by:

$$p_{\text{cascade}}(x_0) = \iint p_1(\hat{x}_0) q(x_N | \hat{x}_0) p_2(x_0 | x_N) d\hat{x}_0 dx_N \quad (B.5)$$

**Lemma 2 (Reconstruction Quality Modeling).** Let the quality of a sampled image be described by a quality function

$$Q(N) = S(N) + D(N) \quad (B.6)$$

where $N$ denotes the number of resample steps, $S(N)$ represents the physical structure consistency, and $D(N)$ is the cross-modal semantic consistency. Mathematically, they are modeled by an exponential form as follows:

$$S(N) = S_1 e^{-k_S N} + S_2 (1 - e^{-k_S N}) \quad (B.7)$$

$$D(N) = D_1 (1 - e^{-k_D N}) + D_2 e^{-k_D N} \quad (B.8)$$

where $S_1$ and $S_2$ denote the physical structure quality scores of $\hat{x}_0$ and $x_0$, respectively. Similarly, $D_1$ and $D_2$ denote the semantic information quality scores of $\hat{x}_0$ and $x_0$, respectively. $k_s$ and $k_d$ denote the score decay factors that vary with $N$.

**Theorem 1 (Existence of an Optimal Resample Timestep).** Consider a resampling process involving SD1 and SD2 over a total of $T$ diffusion steps. Let $N \in [0, T]$ denote the number of resampling timesteps, and let the image quality metric $Q(N)$ be defined as in *Lemma 2*. If the following conditions hold:

(i) SD1 excels over SD2 in structural generation, $S_1 > S_2$.

(ii) SD2 excels over SD1 in semantic generation, $D_2 > D_1$.

(iii) The decay constants satisfy $k_S > 0$, $k_D > 0$, $k_S \neq k_D$.

Then there exists at least one optimal resampling timestep $N^* \in [0, T]$ that maximizes $Q(N)$. Moreover, if the addition condition $\frac{(S_1 - S_2) k_S}{(D_2 - D_1) k_D} > 0$ is satisfied and the quantity $\frac{1}{k_S - k_D} \ln(\frac{(S_1 - S_2) k_S}{(D_2 - D_1) k_D})$ lies in the open interval $(0, T)$, then the optimal solution $N^*$ is given explicitly by

$$N^* = \frac{1}{k_S - k_D} \ln(\frac{(S_1 - S_2) k_S}{(D_2 - D_1) k_D}) \quad (B.9)$$

*Proof.* By *Lemma 1*, the marginal distribution $p_{\text{cascade}}(x_0; N)$ of the cascade sample process is continuous in $N$ because both forward noising and reverse reconstruction depend continuously on the step number. Since the quality measure $S(\cdot)$ and $D(\cdot)$ are assumed to be continuous function, the composite quality function $Q(N) = S(N) + D(N)$ is continuous on the closed interval $[0,T]$. The extreme value theorem therefore guarantees that $Q(N)$ attains a maximum on $[0,T]$, establishing the existence of optimal resampling step.

Using the explicit form in *Lemma 2*, $Q(N)$ can be written as:

$$Q(N) = (S_1 - S_2)e^{-k_s N} + (D_2 - D_1)e^{-k_D N} + (S_2 + D_1) \quad (B.10)$$

Differentiating with respect to $N$ gives:

$$\frac{dQ}{dN} = -(S_1 - S_2)k_s e^{-k_s N} + (D_2 - D_1)k_D e^{-k_D N} \quad (B.11)$$

Setting the derivative to zero and using the assumption $S_1 > S_2$ and $D_2 > D_1$ yields the condition:

$$(D_2 - D_1)k_D e^{-k_D N} = (S_1 - S_2)k_S e^{-k_S N} \quad (B.12)$$

Solving for $N$ by taking logarithms produces a unique point:

$$N^* = \frac{1}{k_S - k_D}\ln(\frac{(S_1 - S_2)k_S}{(D_2 - D_1)k_D}) \quad (B.13)$$

provided $k_S \neq k_D$. Under the conditions $S_1 > S_2$, $D_2 > D_1$ and $k_D > k_S$ (detail information decays faster than structural information), the second derivative test confirms that $N^*$ is a local maximum. If $k_S = k_D$, the function becomes monotonic, and the optimum lies at the boundary. In all cases, the existence of an optimal $N$ is guaranteed. ∎

***Remark 1.*** The optimal $N^*$ balances structural information from SD1 and meta-information semantic from SD2. The $N$ provides the optimal trade-off.

### C. Reverse Process with Multi Conditions

***Theorem 2 (Multi-Condition Guided Reverse Reconstruction).*** Let $lq$ denote a low-dose PET image, $y$ a sinogram, and $m$ meta-information. Assume that three conditions $\{lq, y, m\}$ are conditionally independent given $x_t$. Then the score of the multi-conditional reverse transition can be decomposed as

$$\begin{aligned}&\nabla_{x_t}\log p(x_t|x_{t+1},lq,y,m)\\&=\nabla_{x_t}\log p(x_t|x_{t+1})+\sum_{c\in\{lq,y,m\}}\lambda_c\nabla_{x_t}\log p(c|x_t)\end{aligned} \quad (C.1)$$

where $\lambda_c > 0$ are scaling coefficients that control the relative strength of guidance from each condition.

***Proof.*** By *Bayes' theorem*, the multi-conditional reverse transition satisfies

$$p(x_t|x_{t+1},lq,y,m) = \frac{p(x_{t+1},lq,y,m|x_t)p(x_t)}{p(x_{t+1},lq,y,m)} \quad (C.2)$$

Taking the gradient with respect to $x_t$ of the logarithmic of both sides yields

$$\begin{aligned}&\nabla_{x_t}\log p(x_t|x_{t+1},lq,y,m)=\\&\nabla_{x_t}\log p(x_{t+1},lq,y,m|x_t)+\nabla_{x_t}\log p(x_t)\end{aligned} \quad (C.3)$$

Under the conditional-independence assumption, the joint likelihood factorizes as

$$\begin{aligned}&p(x_{t+1},lq,y,m|x_t)=\\&p(x_{t+1}|x_t)p(lq|x_t)p(y|x_t)p(m|x_t)\end{aligned} \quad (C.4)$$

Substituting Eq. (C.3) into Eq. (C.4) gives

$$\begin{aligned}&\nabla_{x_t}\log p(x_t|x_{t+1},lq,y,m)=\nabla_{x_t}\log p(x_t)+\\&\nabla_{x_t}\log p(x_{t+1}|x_t)+\sum_{c\in\{lq,y,m\}}\nabla_{x_t}\log p(c|x_t)\end{aligned} \quad (C.5)$$

Applying Bayes' theorem to the first two terms on the right-hand side,

$$\nabla_{x_t}\log p(x_t)+\nabla_{x_t}\log p(x_{t+1}|x_t)=\nabla_{x_t}\log p(x_t|x_{t+1}) \quad (C.6)$$

Combining Eq. (C.5) and Eq. (C.6), we obtain

$$\begin{aligned}&\nabla_{x_t}\log p(x_t|x_{t+1},lq,y,m)\\&=\nabla_{x_t}\log p(x_t|x_{t+1})+\sum_{c\in\{lq,y,m\}}\nabla_{x_t}\log p(c|x_t)\end{aligned} \quad (C.7)$$

Finally, introducing the scaling coefficients $\lambda_c$ to balance the influence of each conditional term yields the desired expression Eq. (C.1). ∎

***Remark 2.*** Setting $\lambda_c = 0$ for all $c \in \{lq, y, m\}$ in Eq. (C.1) recovers the unconditional score $\nabla_{x_t}\log p(x_t|x_{t+1})$. Consequently, the multi-condition guided reverse process can be decomposed into three mutually independent gradients injected into an unconditional diffusion model. In the MiG-DM architecture, the image domain condition $lq$ is furnished by the SD1; the projection-domain condition $y$ is provided by the SinoA; and the Meta-information condition $m$ is generated by the MI-encoder.